\documentclass[10pt,twocolumn,letterpaper]{article}

\usepackage{iccv}
\usepackage{times}
\usepackage{epsfig}
\usepackage{graphicx}
\usepackage{amsmath}
\usepackage{amssymb}

\newcommand{\bs}[1]{\boldsymbol{#1}}

\DeclareMathOperator*{\argmin}{arg~min}
\usepackage{comment}

\usepackage{booktabs}       

\usepackage[ruled,vlined,noend]{algorithm2e}

\usepackage{soul}
\usepackage{marvosym}
\newcommand\markUnbabel{\Cancer}
\newcommand\markIT{\Leo}
\newcommand\markISR{\Jupiter}
\newcommand\markLUMLIS{\Saturn}

\usepackage[pagebackref=true,breaklinks=true,letterpaper=true,colorlinks,bookmarks=false]{hyperref}

\iccvfinalcopy

\def\iccvPaperID{***} 

\ificcvfinal\fi

\begin{document}

\title{Multimodal Continuous Visual Attention Mechanisms}

\author{%
  Ant\'onio Farinhas\textsuperscript{\markIT}
 \and
  Andr\'e F.~T.~Martins\textsuperscript{\markIT,\markLUMLIS,\markUnbabel} 
  \and
  Pedro M.~Q.~Aguiar\textsuperscript{\markISR,\markLUMLIS} 
  \and \\[-2ex]
  \{
  {\tt\small antonio.farinhas}, 
  {\tt\small andre.t.martins}\}{\tt\small @tecnico.ulisboa.pt},
  {\tt\small aguiar@isr.ist.utl.pt} \\[1ex]
  \textsuperscript{\markIT{}}Instituto de Telecomunica\c{c}\~oes,  Instituto Superior Técnico, Lisbon, Portugal \\
  \textsuperscript{\markISR{}}Instituto de Sistemas e Rob\'otica, Instituto Superior Técnico, Lisbon, Portugal \\
  \textsuperscript{\markLUMLIS{}}LUMLIS (Lisbon ELLIS Unit), Lisbon, Portugal \\
  \textsuperscript{\markUnbabel{}}Unbabel, Lisbon, Portugal
}

\maketitle
\ificcvfinal\thispagestyle{empty}\fi

\begin{abstract}
   Visual attention mechanisms are a key component of neural network models for computer vision.
   By focusing on a discrete set of objects or image regions, these mechanisms identify the most relevant features and use them to build more powerful representations.  
   Recently, continuous-domain alternatives to discrete attention models have been proposed, which exploit the continuity of images.
   These approaches model attention as simple unimodal densities (\eg a Gaussian), making them less suitable to deal with images whose region of interest has a complex shape or is composed of multiple non-contiguous patches.
   
   In this paper, we introduce a new continuous attention mechanism that produces multimodal densities, in the form of mixtures of Gaussians.
   We use the EM algorithm to obtain a clustering of relevant regions in the image, and a description length penalty to select the number of components in the mixture.
   Our densities decompose as a linear combination of unimodal attention mechanisms, enabling closed-form Jacobians for the backpropagation step.
   Experiments on visual question answering in the VQA-v2 dataset show competitive accuracies and a selection of regions that mimics human attention more closely in VQA-HAT.
   We present several examples that suggest how multimodal attention maps are naturally more interpretable than their unimodal counterparts, showing the ability of our model to automatically segregate objects from ground in complex scenes.

\end{abstract}

\section{Introduction}

\textbf{Visual attention mechanisms} are an important component of modern deep learning models \cite{Xu2015-show-attend-tell-captioning, Andreas2016-neural-module-nets, sharma2015actrecNIPS, Zhang_2018_CVPR}.
They appear as a way to mimic the human visual system, which selectively attends to the most relevant parts of visual stimuli, enabling processing large amounts of
information in parallel \cite{Rensink-2000-visual}.

A neural network with an attention mechanism automatically learns the relevance of any element of the input by generating a set of weights and taking them into account while performing the proposed task. 
In addition to boosting the performance of a model, attention mechanisms can provide insights into the model's decision process, being suitable for interpretability purposes \cite{wiegreffe-pinter-2019-attention-not-not-explanation, vqahat}.
In particular, the visualization of attention weights can help us analyze the outputs of a neural network and possibly understand some unpredictable outcomes \cite{Galassi2020-att-review}.

\begin{figure*}[t]
\centering
\includegraphics[width=0.33\linewidth]{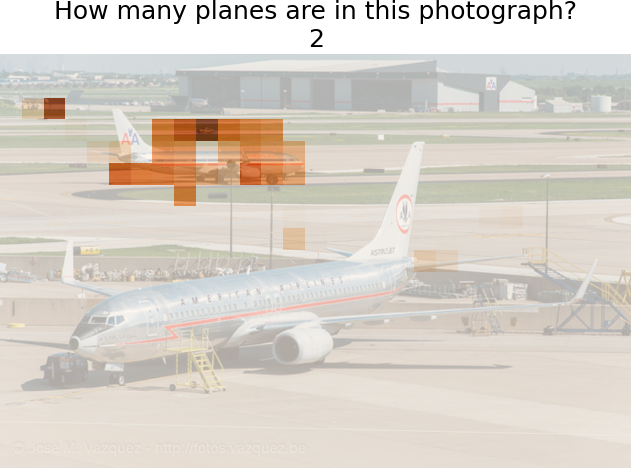}
\includegraphics[width=0.33\linewidth]{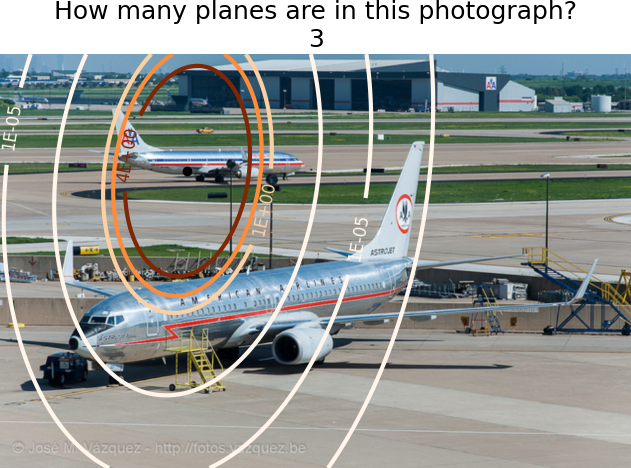}
\includegraphics[width=0.33\linewidth]{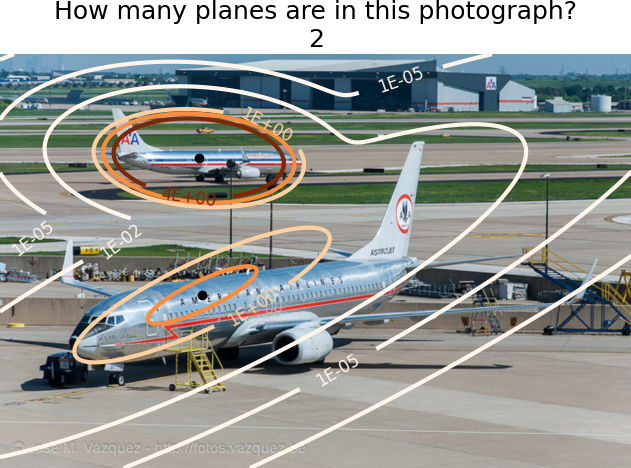}
\caption{\label{fig:uni-vs-multimodal} Examples of attention maps for VQA. Left: discrete attention. Middle: Unimodal continuous attention. Right: Multimodal continuous attention (ours). For continuous attention models, we identify the means of the Gaussians with black dots.
}
\end{figure*}

Most models for visual attention operate over discrete domains, where images are split into a finite set of regions or pixels \cite{Xu2015-show-attend-tell-captioning, Andreas2016-neural-module-nets, sharma2015actrecNIPS, Zhang_2018_CVPR}. 
However, this sometimes leads to lack of focus, where the attention distribution over the image becomes too scattered. 
Discrete attention mechanisms disregard the fact that images are inherently ``continuous'' objects.  
Recently, \textbf{continuous attention mechanisms} have been proposed \cite{ContinuousAttention2020}, which are able to attend over continuous domains and to select compact regions of interest in the image, such as ellipses. 
Nevertheless, this approach (which we review in \S\ref{section:continuous}) is limited in which it models attention with a simple unimodal density, making it less suitable to deal with images whose region of interest has a complex shape or is composed of multiple non-contiguous patches.

In this paper, we address the limitation above by introducing \textbf{multimodal} continuous attention mechanisms, in the form of mixtures of unimodal distributions (\S\ref{section: Multimodal continuous attention}). 
These mechanisms are able to generate more flexible attention maps while enjoying the best properties of their unimodal counterparts. 
In particular, we study the case where the attention density is modeled by a mixture of Gaussians. We use the Expectation-Maximization (EM) algorithm to obtain a clustering of relevant regions in the image (\S\ref{section: The EM algorithm for GMMs}), and we apply a description length penalty to select the number of components in the mixture (\S\ref{section: Estimating the number of components}).
Crucially, our densities decompose as a linear combination of unimodal attention mechanisms, enabling tractable and efficient forward and gradient  backpropagation steps.

Our experiments in visual question answering show competitive accuracy results in the VQA-v2 dataset~\cite{balanced_vqa_v2} (\S\ref{section:Experiments}). More compelling is the fact that the proposed models lead to more interpretable decisions, being able, for example, to
attend to multiple objects without becoming overly unfocused, as illustrated by the example in Figure~\ref{fig:uni-vs-multimodal}. To obtain a quantitative measure of how well artificial models represent human attention, we use the VQA-HAT dataset~\cite{vqahat}, concluding that the attention maps provided by the proposed models lead to an overall higher similarity than the ones obtained with discrete or unimodal continuous attention.

\section{Continuous attention}
\label{section:continuous}

\subsection{Discrete attention}

Attention mechanisms are typically discrete \cite{bahdanau2014neural,Xu2015-show-attend-tell-captioning}. In vision applications, the starting point is an input image from which $L$ feature vectors in $\mathbb{R}^D$ are extracted (e.g., grid-level or object-level representations), leading to a feature matrix $\bs{V} \in \mathbb{R}^{D \times L}$. 
Given some conditioning context (for example a question in natural language), a \textit{score vector} $\bs{f}=[f_1,\ldots,f_L]^\top \in \mathbb{R}^L$ is computed,
where high scores correspond to more relevant parts of the input. 
These scores are converted into a probability vector $\bs{p} \in \bigtriangleup^L$ (the \textit{attention weights}), where $\bigtriangleup^L := \{\bs{p}\in\mathbb{R}^L \mid \mathbf{1}^\top \bs{p} = 1, \bs{p} \geq \mathbf{0}\}$ is the $L$-dimensional probability simplex,  
typically via a softmax transformation, $\bs{p} = \mathrm{softmax}(\bs{f})$. 
Finally, the probability vector is used to compute a weighted average of the input (known as the \textit{context vector}), $\bs{c}=\bs{V}\bs{p} \in \mathbb{R}^D$, 
that is used to produce the network's decision (for example, an answer to the question).

While discrete attention mechanisms are very flexible, since they allow arbitrary probability mass functions over the input features, this flexibility can be harmful, resulting sometimes in attention maps that are too scattered and lack focus -- this may affect prediction accuracy and result in poor interpretability.

\subsection{Continuous attention}
\label{subsection:Continuous attention}

To avoid the shortcoming above, Martins~\etal\cite{ContinuousAttention2020} introduced \textit{continuous} attention mechanisms, where images are represented as smooth functions in 2D, instead of being split into regions in a grid. 

\paragraph{Feature function.} In this framework, instead of the feature matrix $\bs{V} \in \mathbb{R}^{D \times L}$ above, the image is represented as a continuous \textit{feature function} $\bs{V}: \mathbb{R}^2 \rightarrow \mathbb{R}^D$, where each point in the $\mathbb{R}^2$ plane is assigned a vector representation. This function is linearly parametrized as 
\begin{equation}\label{eq:value_function}
\bs{V}_{\!\!\bs{B}}(\bs{x})=\bs{B}\bs{\psi}(\bs{x}),
\end{equation}
where $\bs{x} = [u,v]^\top$ are coordinates in the image,  $\bs{\psi}: \mathbb{R}^2 \rightarrow \mathbb{R}^N$ are $N$ bivariate Gaussian radial basis functions (RBFs) with different means and covariance parameters, and $\bs{B} \in \mathbb{R}^{D \times N}$ are parameters fit with ridge regression (see~\cite[\S3.1]{ContinuousAttention2020} for details).
If $N \ll L$ (fewer basis functions than regions), the continuous representation of the image is more compact than the discrete feature matrix.

\paragraph{Score function and attention density.} Likewise, the score vector $\bs{f}=[f_1,\ldots,f_L]^\top$ above is replaced by a quadratic \textit{score function} $f: \mathbb{R}^2 \rightarrow \mathbb{R}$, defined as 
\begin{equation}\label{eq:score_function}
f(\bs{x})=-\frac{1}{2}(\bs{x}-\bs{\mu})^\top \bs{\Sigma}^{-1}(\bs{x}-\bs{\mu}),
\end{equation}
where $\bs{\mu} \in \mathbb{R}^2$ is a location parameter and $\bs{\Sigma} \succ \mathbf{0}$ is a positive definite matrix in $\mathbb{R}^{2\times 2}$. 
This way, relevance is directed to a single location in the image (specified by $\bs{\mu}$) and it has an elliptical shape, determined by $\bs{\Sigma}$. 
The score function is mapped to a probability density $p: \mathbb{R}^2 \rightarrow \mathbb{R}_+$ via a regularized prediction mapping \cite{blondel2020learning}. With an entropy regularizer, this results in a Gibbs distribution $p(\bs{x}) \propto \exp(f(\bs{x}))$, which for quadratic scores leads to a Gaussian density, $p(\bs{x}) = \mathcal{N}(\bs{x}; \bs{\mu}, \bs{\Sigma})$.

\paragraph{Evaluation and gradient backpropagation.} In continuous attention mechanisms, the output weighted average (context vector) is written as the  expectation of the feature function with respect to the probability density, 
\begin{equation}\label{eq:context_vector}
\bs{c}=\mathbb{E}_p[\bs{V}_{\!\!\bs{B}}(\bs{x})] = \bs{B} \int_{\mathbb{R}^2} p(\bs{x})\bs{\psi}(\bs{x}) \in \mathbb{R}^D,
\end{equation}
where we used \eqref{eq:value_function}. 
If $\bs{\psi}(\bs{x})$ are Gaussian RBFs and $p(\bs{x})$ is a Gaussian, expression \eqref{eq:context_vector} becomes the integral of a product of Gaussians, which has a closed form. 
The backpropagation step can be done either with automatic differentiation or by using a covariance expression to compute the Jacobians ${\partial \bs{c}}/{\partial \bs{\mu}}$ and ${\partial \bs{c}}/{\partial \bs{\Sigma}}$ \cite[\S3.2]{ContinuousAttention2020}.

\section{Multimodal continuous attention}
\label{section: Multimodal continuous attention}

In this paper, we extend the continuous attention framework described in \S\ref{subsection:Continuous attention} to \textbf{multimodal distributions}. This is done by letting the attention density be a mixture of unimodal distributions $p_k: \mathbb{R}^2 \rightarrow \mathbb{R}_+$, for $k \in \{1, \ldots, K\}$:
\begin{equation}
\label{eq:multimodal-attention-prob}
    p(\bs{x}) = \sum_{k=1}^{K} \pi_k  p_k(\bs{x}),
\end{equation} 
where 
$\bs{\pi} = [\pi_1, \ldots, \pi_K]^\top \in \Delta^K$
are mixing coefficients, defining the weight of each component of the mixture. We let each $p_k(\bs{x})$ be a Gaussian distribution, so that $p(\bs{x})$ becomes a mixture of Gaussians; we discuss below possible methods for obtaining the mixing coefficients $\bs{\pi}$. 
By doing this, we increase the expressive power of continuous attention mechanisms: by using a sufficient number of Gaussians and adjusting the parameters of the mixture, almost any continuous density can be approximated to arbitrary accuracy \cite[\S2.3.9]{bishop}.

\paragraph{Forward step.}
Using \eqref{eq:context_vector} and invoking the linearity of expectations, we can compute the output of the multimodal attention mechanism as
\begin{equation}\label{eq:multimodal_context}
    \bs{c} = \mathbb{E}_p[\bs{B}\bs{\psi}(\bs{x})] = \sum_{k=1}^K \pi_k \underbrace{\mathbb{E}_{p_k}[\bs{B}\bs{\psi}(\bs{x})]}_{\bs{c}_k} = \sum_{k=1}^K \pi_k \bs{c}_k,
\end{equation} 
where each $\bs{c}_k$ is the context representation after applying each individual (unimodal) attention mechanism; \ie, \textbf{$\bs{c}$ is a mixture of the context representations for each component}. 

\paragraph{Backpropagation step.}
The backpropagation step for the multimodal case is also simple, since it decomposes into a linear combination of unimodal attention mechanisms, each of which has a simple/closed-form Jacobian.

\paragraph{Relation to multi-head attention.} 
Our multimodal attention has some resemblances with multi-head attention mechanisms \cite{Vaswani2017}, if we regard each component of the mixture as if it were a different attention head.
Note, however, that our construction differs from multi-head attention, where the projection matrices learned as model parameters are head-specific. On the contrary, we assume that $\bs{B}$ in \eqref{eq:multimodal_context} is fixed, \ie, it does not depend on $k$. This avoids head-specific computations and enables a probabilistic interpretation of the resulting density as a mixture of densities.

\paragraph{How can we estimate the parameters of the attention density?}
To choose the mixing coefficients $\bs{\pi}$, along with the means $(\bs{\mu}_k)_{k=1}^K$ and covariance matrices $(\bs{\Sigma}_k)_{k=1}^K$ of each component of the mixture, we start from a given set of observed image locations along with their importance weights $\{(\bs{x}_{\ell}, w_\ell)\}_{\ell=1}^L$. 
Intuitively, the higher the weight, the more important the contribution of that specific region should be to the network's decision.
To parametrize the attention density as a simple unimodal distribution, it is possible to use moment matching.
For multimodal distributions, we can think of this problem as that of fitting a mixture model to weighted data. In that context, we have to deal with two different issues: how to estimate the number of components, which we discuss in \S\ref{section: Estimating the number of components}, and how to estimate the parameters defining the mixture model. A popular choice to address the second problem is the EM algorithm, which seeks a maximum likelihood estimate of the mixture parameters and is guaranteed to converge to a local maximum  \cite{Dempster77maximumlikelihood, mclachlan-1997-EM}.
If $p_k$ is a Gaussian, we can easily adapt the EM algorithm to deal with weighted data (\eg, discrete attention weights and corresponding grid locations), so that we can estimate the full set of parameters of a mixture of Gaussians -- defining a multimodal attention density, $p(\bs{x})$. 
This is described in detail in the next section.

\section{The EM algorithm for mixtures of Gaussians}
\label{section: The EM algorithm for GMMs}
The EM algorithm is the standard method to estimate the parameters defining a mixture model. It starts with an initial estimate for the parameters of the mixture and iteratively updates them until convergence or up to a predefined number of iterations
(see \cite{mclachlan-1997-EM} for a detailed exposition).
In this paper, we assume that each component of the mixture is a Gaussian, \ie, the multimodal attention density $p(\bs{x})$ takes the form of a mixture of Gaussians.

\subsection{EM with weighted data}
\label{subsubsection: Weighted data}
Let $\mathcal{X} = \{(\bs{x}_1, w_1), \ldots, (\bs{x}_L, w_L)\}$ be the observed data along with their weights. In our approach, each $\bs{x}_\ell \in \mathbb{R}^2$ is the center of a grid region, and $w_\ell \in [0,1]$ is the corresponding discrete attention weight.
Our goal is to maximize the likelihood function
\begin{equation}
\mathcal{L}(\Theta) = \sum_{\ell=1}^L w_\ell \log p(\bs{x}_\ell \mid \Theta),
\end{equation}
where $\Theta = \{(\pi_k, \bs{\mu}_k, \bs{\Sigma}_k)\}_{k=1}^K$. 
We adapt the  EM algorithm for mixtures of Gaussians to handle weighted data, by changing the way the parameters are  re-estimated at each iteration. The algorithm  goes as follows:
\begin{enumerate}
    \item Initialize the parameters $\{\pi_k, \bs{\mu}_k, \bs{\Sigma}_k\}_{k=1}^K$ and evaluate the initial value of the weighted log-likelihood function:
    \begin{equation}
        \mathcal{L}(\Theta) = \sum_{\ell=1}^L w_\ell\log \left\{\sum_{k=1}^K \pi_k \mathcal{N}(\bs{x}_\ell; \bs{\mu}_k, \bs{\Sigma}_k) \right\},
    \label{eq:EM-wgt-loglikelihood}
    \end{equation}
    where the log-likelihood of each point is multiplied by the correspondent weight.

    \item \textbf{E step.} Evaluate the responsibilities using the current parameter values: 
    \begin{equation}
        \gamma_{\ell k} = \frac{\pi_k \mathcal{N}(\bs{x}_\ell| \bs{\mu}_k, \bs{\Sigma}_k)}{\sum_{j=1}^{K} \pi_j \mathcal{N}(\bs{x}_\ell| \bs{\mu}_j, \bs{\Sigma}_j)}.
    \label{eq:EM-responsabilities}
    \end{equation}
    
    \item\textbf{M step.} Re-estimate the parameters using the current responsibilities:
    \begin{equation}
        \pi_k^{\mathrm{new}} = \sum_{\ell=1}^L w_\ell\gamma_{\ell k},
    \label{eq:WD-EM-pi-k-new}
    \end{equation}  
    \begin{equation}
        \bs{\mu}_k^{\mathrm{new}} = \frac{1}{\pi_k^{\mathrm{new}}} \sum_{\ell=1}^{L} w_\ell \gamma_{\ell k}\bs{x}_\ell,
    \label{eq:WD-EM-mu-k-new}
    \end{equation}
    \begin{equation}
        \bs{\Sigma}_k^{\mathrm{new}} = \frac{1}{\pi_k^{\mathrm{new}}}\sum_{\ell=1}^{L} w_\ell \gamma_{\ell k}(\bs{x}_\ell - \bs{\mu}_k^{\mathrm{new}})(\bs{x}_\ell - \bs{\mu}_k^{\mathrm{new}})^{\top}.
    \label{eq:WD-EM-Sigma-k-new}
    \end{equation}    
        
    \item Re-evaluate the weighted log-likelihood \eqref{eq:EM-wgt-loglikelihood} using the current parameter values and check for convergence of either the parameters or the log likelihood. Return to step 2 if the convergence criterion is not satisfied.
    
\end{enumerate}
If the weight associated with each observation is the same, \textit{i.e.}, $w_\ell = {1}/{L}$, we recover the usual expressions for the EM algorithm.

\subsection{Initialization}
\label{subsection: Initialization}
The EM algorithm requires an initial choice for the set of parameters $\Theta = \{\pi_k, \bs{\mu}_k, \bs{\Sigma}_k \}_{k=1}^K$. 
This is a relevant issue because EM is not guaranteed to converge to a global maximizer of the log-likelihood function, but rather a local one,
meaning that the final estimate depends on the initialization.
An effective strategy is to run EM multiple times with different random initializations and choose the final estimate that leads to the highest likelihood \cite{mclachlan-1997-EM}.

\section{Estimating the number of components}
\label{section: Estimating the number of components}
The maximum likelihood criterion cannot be used to estimate the number of components $K$ in a mixture density: If $\mathcal{M}_k$ is a class composed by all Gaussian mixtures with $K$ components, it is trivial to show that $\mathcal{M}_K \subseteq \mathcal{M}_{K+1}$ and thus the maximized likelihood is a non decreasing function of $K$, useless as a criterion to estimate $K$ 
\cite{Figueiredo00unsupervisedlearning}. 
For this reason, 
several \textbf{model selection} methods have been proposed to estimate the number of components of a mixture \cite[Chapter~6]{McLachlan2000-finitemixmodels}.
We focus on penalized likelihood methods such as the Bayesian Information Criterion (BIC, \cite{Schwarz-1978-BIC}) or the Minimum Description Length (MDL, \cite{Rissanen-1989-MDL}), where the EM algorithm is used to obtain different parameter estimates for a range of values of $k$, $\{\hat{\Theta}_k,~k=k_{\mathrm{min}},\dots,k_{\mathrm{max}}\}$, and the number of components is chosen according to 
\begin{equation}
\label{eq:k-argmin}
    k^{\star} = \argmin_{k \in \{k_{\mathrm{min}}, \ldots, k_{\mathrm{max}}\}}\mathcal{C}(\hat{\Theta}_k, k),
\end{equation} where $\mathcal{C}(\hat{\Theta}_k, k)$ is a model selection criterion. We use a criterion of the form
\begin{equation}
    \mathcal{C}(\hat{\Theta}_k,k) = - 2\log{p(\mathcal{X}|\hat{\Theta}_k)} + \mathcal{P}(k),
\end{equation} where $\mathcal{P}(k)$ is an increasing function penalizing higher values of $k$ (\eg, $\mathcal{P}_{\mathrm{BIC}}(k) = k \log{n}$, where $n$ is the number of data points). 
For the weighted data scenario presented in \S\ref{subsubsection: Weighted data} we cannot use the number of points; thus, we  write
\begin{equation}
    \mathcal{P}(k) = \lambda~k,
\end{equation} where $\lambda > 0$ is an hyperparameter obtained using \textit{cross-validation}. The resulting model selection criterion,
\begin{equation}
\label{eq:model-selection-criterion-lambda}
    \mathcal{C}(\hat{\Theta}_k,k) = - 2\log{p(\mathcal{X}|\hat{\Theta}_k)} + \lambda~k,
\end{equation} will be used in \S\ref{section:Experiments} to estimate the number of components in a multimodal continuous attention density.

\begin{algorithm*}[ht]
\small
\SetAlgoLined
\SetKwInput{KwInput}{Parameters}
\SetKwFunction{FRegression}{Regression}
\SetKwFunction{FForward}{MultimodalAttention}
\SetKwFunction{FBackward}{Backward-unimodal}
\SetKwFunction{FEM}{WeightedEM}
\SetKwFunction{FSelection}{ModelSelection}
\def\algspace{.5\baselineskip}
\SetKwProg{Fn}{Function}{:}{}
\KwInput{Centers of grid regions and their weights $\mathcal{X} = \{(\bs{x}_\ell, w_\ell)\}_{\ell=1}^L$, initialization $\Theta(K) = \{(\pi_k, \bs{\mu}_k, \bs{\Sigma}_k)\}_{k=1}^{K}$, number of iterations $I$, Gaussian RBFs $\bs{\psi}(\bs{x}) = [\mathcal{N}(\bs{x}; \bs{\mu}_j, \bs{\Sigma}_j)]_{j=1}^N$, value function $\bs{V}_{\!\!\bs{B}}(\bs{x}) = \bs{B}\bs{\psi}(\bs{x})$.}

\vspace{\algspace}
\Fn{\FEM{$\mathcal{X}, \Theta(K), I$}}{

    \For{$i\gets1$ \KwTo I}{
    
        \For{$\ell \gets1$ \KwTo L}{
            \For{$k\gets1$ \KwTo $K$}{
                $\gamma_{\ell k} \leftarrow \frac{\pi_k \mathcal{N}(\bs{x}_\ell| \bs{\mu}_k, \bs{\Sigma}_k)}{\sum_{j=1}^{K} \pi_j \mathcal{N}(\bs{x}_\ell| \bs{\mu}_j, \bs{\Sigma}_j)}$ \hfill%
                \tcp*[h]{\eqref{eq:EM-responsabilities}}\\
            }
        
        }

        \For{$k\gets1$ \KwTo $K$}{
            $\pi_k \leftarrow \sum_{\ell=1}^L w_l\gamma_{\ell k}$\hfill%
                \tcp*[h]{\eqref{eq:WD-EM-pi-k-new}}\\ 
            $\bs{\mu}_k \leftarrow \frac{1}{\pi_k} \sum_{\ell=1}^{L} w_\ell \gamma_{\ell k}\bs{x}_\ell$, \,\,
            $\bs{\Sigma}_k \leftarrow \frac{1}{\pi_k}\sum_{\ell=1}^{L} w_\ell \gamma_{\ell k}(\bs{x}_\ell - \bs{\mu}_k)(\bs{x}_\ell - \bs{\mu}_k)^{\top} $ \hfill%
                \tcp*[h]{\eqref{eq:WD-EM-mu-k-new}, \eqref{eq:WD-EM-Sigma-k-new}}\\ 
        
        }

    }
    \KwRet{$\Theta = \{(\pi_k, \bs{\mu}_k, \bs{\Sigma}_k)\}_{k=1}^{K}$}

}

\vspace{\algspace}
\Fn{\FSelection{$\mathcal{X}, \{\Theta(k)\}_{k=1}^{k_{\mathrm{max}}}, I, \lambda$}}{

    \For{$k\gets1$ \KwTo $k_{\mathrm{max}}$}{
    
        $\hat{\Theta}_k \leftarrow$ \FEM{$\mathcal{X},\Theta(k),I$} \\
        $ \log{p(\mathcal{X}|\hat{\Theta}_k)} \leftarrow \sum_{\ell=1}^L w_\ell\log \left\{\sum_{k=1}^K \hat{\pi}_k \mathcal{N}(\bs{x}_\ell|\hat{\bs{\mu}}_k, \hat{\bs{\Sigma}}_k) \right\}$, \,\,
        $\mathcal{C}(\hat{\Theta}_k,k) \leftarrow - 2\log{p(\mathcal{X}|\hat{\Theta}_k)} + \lambda~k$ \hfill%
        \tcp*[h]{\eqref{eq:EM-wgt-loglikelihood}, \eqref{eq:model-selection-criterion-lambda}}\\ 
    
    }
    
    $k^{\star} = \argmin_k\{\mathcal{C}(\hat{\Theta}_k, k)\}$ \hfill%
                \tcp*[h]{\eqref{eq:k-argmin}}\\
    \KwRet{$k^\star, \hat{\Theta}_{k^\star}$}
}

\vspace{\algspace}
\Fn{\FForward{$\bs{V}_{\!\!\bs{B}}$, $\Theta = \{(\pi_k, \bs{\mu}_k, \bs{\Sigma}_k)\}_{k=1}^{K}$}}{

    \For{$k\gets1$ \KwTo $K$}{
        $r_{kj} \leftarrow \mathbb{E}_p[\psi_j(\bs{x})] = 
            \mathcal{N}(\bs{\mu}_k, \bs{\mu}_j, \bs{\Sigma}_k+\bs{\Sigma}_j), \quad \forall j \in [N]$ \hfill%
            \tcp*[h]{\cite[\S3]{ContinuousAttention2020}}\\
        $\bs{c}_k \leftarrow \bs{B}\bs{r}_k$ \hfill%
        \tcp*[h]{\eqref{eq:context_vector}}\\
    }
    \vspace{\algspace}
    
    $\bs{c} \leftarrow \sum_{k=1}^K \pi_k \bs{c}_k$ \hfill%
    \tcp*[h]{\eqref{eq:multimodal_context}}\\

    \KwRet{$\bs{c}$ (context vector)}%
}
\vspace{\algspace}
\vspace{\algspace}

\caption{Multimodal continuous attention with Gaussian RBFs. During training, we pick the number of components randomly and apply $\mathrm{WeightedEM}$, followed by $\mathrm{MultimodalAttention}$. At test time, we apply $\mathrm{ModelSelection}$ in between the previous functions to select the number of components.
\label{algo:multimodal_attention}}
\end{algorithm*}

\paragraph{Attention model.} 
Using the results of the previous section, we model each attention density as a $K$-component mixture of Gaussians.
At training time, we pick the number of components randomly from a uniform distribution, up to a predefined maximum. This way we expose the model to different numbers of components, maintaining the simplicity of the training procedure without added runtime. 
At test time, we select the optimum $K^\star$ from a set of possible choices, using the model selection criterion \eqref{eq:model-selection-criterion-lambda}.
See Algorithm~\ref{algo:multimodal_attention} for pseudo-code. 
(Although we consider multiple random initializations along with the model selection criterion, we omit this step in the algorithm, for simplicity.)

Note that our extension from unimodal to multimodal continuous attention does not increase the number of neural network parameters. Therefore, in practice, it is possible to leverage the learned representations from a pretrained model using either discrete or unimodal continuous attention mechanisms (\eg discrete or continuous softmax) and fine-tune it with our multimodal attention densities. This method allows us to model the attention distribution as an expressive density function that could not be properly modeled using a single Gaussian.

\section{Experiments}
\label{section:Experiments}

\subsection{Visual Question Answering}

\begin{table*}[ht]
    \begin{center}
\begin{tabular}{l@{\hspace{15pt}}c@{\hspace{5pt}}c@{\hspace{5pt}}c@{\hspace{5pt}}c@{\hspace{15pt}}c@{\hspace{5pt}}c@{\hspace{5pt}}c@{\hspace{5pt}}cc}
        \toprule
        \sc Attention & \multicolumn{4}{c}{Test-Dev}  & \multicolumn{4}{c}{Test-Standard} \\
        {} & Yes/No & Number & Other & Overall & Yes/No & Number & Other & Overall \\
        \midrule 
        Discrete softmax & 86.76 & 52.90 & 60.78 & 70.59 & 86.91 & 53.22 & 61.10 & 70.94 \\
        \midrule
        Unimodal continuous & 86.57 & 53.69 & 60.38 & 70.41 & 86.73 & 53.55 & 60.75 & 70.73 \\
        Multimodal continuous & 86.62 & 53.23 & 60.46 & 70.42 & 86.88 & 53.31 & 60.79 & 70.79 \\ 
        \bottomrule
    \end{tabular}
\end{center}
\caption{Accuracies of different models on the \textit{test-dev} and \textit{test-standard} splits of VQA-v2.
}
\label{table:results_vqa}
\end{table*}

\begin{figure*}[t]
\centering
\includegraphics[width=0.33\linewidth]{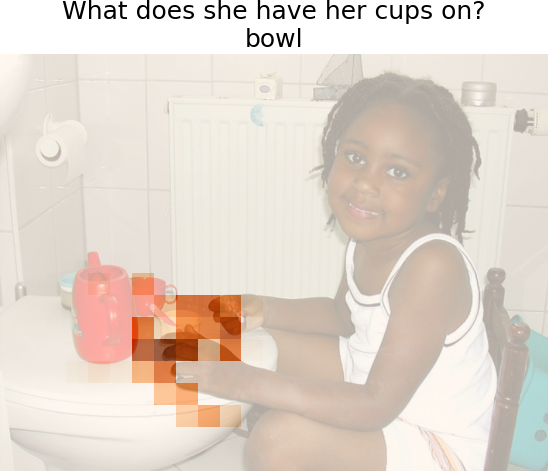}
\includegraphics[width=0.33\linewidth]{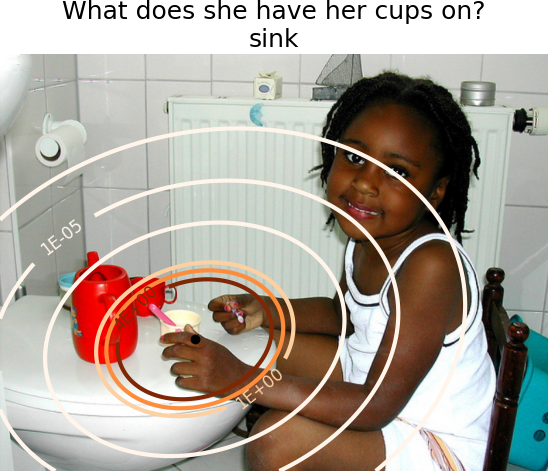}
\includegraphics[width=0.33\linewidth]{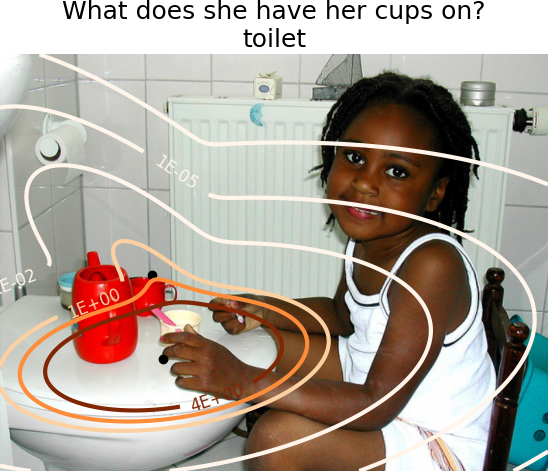}
\caption{\label{fig:toilet} Examples of attention maps in the VQA-v2 dataset. Left: discrete softmax attention. Middle: unimodal continuous attention. Right: Multimodal continuous attention (ours).
}
\end{figure*}

\paragraph{Dataset and metrics.} We use the VQA-v2 dataset \cite{balanced_vqa_v2} with the standard splits (443K, 214K, and 453K question-image pairs for train/dev/test, the latter subdivided into test-dev, test-standard, test-challenge and test-reserve). We report results in terms of accuracy in the test-dev and test-standard splits.
All the models we experiment with are trained only on the train split, without data augmentation.

\paragraph{Architecture.} We adapt the implementation of the encoder-decoder version of the Modular Co-Attention Network (MCAN, \cite{Yu2019}),%
\footnote{\url{https://github.com/MILVLG/mcan-vqa}} %
and represent the image input with grid features generated by Jiang \etal\cite{grid-features-vqa}, using a ResNet pretrained on Visual Genome \cite{visual-genome} that outputs a feature map of size $L \times 2048$, where $L$ is the number of features ($\overline{L} = 506$ and $L_{\mathrm{max}} = 608$). To represent the question words we use 300-dimensional GloVe word embeddings~\cite{pennington2014glove}, yielding a feature matrix representation.

\paragraph{Attention models.}
We consider three different attention models: discrete attention, unimodal continuous  attention, and multimodal continuous attention (ours).
The discrete attention model attends over a grid and uses the softmax transformation to map scores into probabilities.
For the continuous attention models, we normalize the image size into the unit square $[0,1]^2$. Then, we transform the image into a continuous function $\bs{V}_{\!\!\bs{B}} : \mathbb{R}^2 \rightarrow \mathbb{R}^D$ using ridge regression, and fit a Gaussian (unimodal continuous attention) or a mixture of Gaussians (multimodal continuous attention) as the attention density. 
In the first case, we obtain $\bs{\mu}$ and $\bs{\Sigma}$ with moment matching;
in the second case, we use the method described in \S\ref{section: Multimodal continuous attention} -- we set the maximum number of components to $k_{\mathrm{max}} = 4$ and, during training, we pick the number of components randomly from a uniform distribution; at test time, we use $3$ random initializations for each $k$ and apply the model selection criterion \eqref{eq:model-selection-criterion-lambda} to choose the optimum number of components.
In both cases, we use $N=100 \ll 506$ Gaussian RBFs $\mathcal{N}(\bs{x};\Tilde{\bs{\mu}}, \Tilde{\bs{\Sigma}})$, with $\tilde{\bs{\mu}}$ linearly spaced in  $[0,1]^2$ and $\tilde{\bs{\Sigma}}=0.001\cdot \bs{\mathrm{I}}$.
The number of neural network parameters is the same in all attention models, both discrete and continuous.

\paragraph{Settings.}
All models are trained for a maximum of $15$ epochs using the Adam optimizer \cite{adam-optimizer} with a learning rate of $\mathrm{min}(2.5 t \cdot 10^{-5}, 5\cdot 10^{-4})$, where $t$ is the epoch number. After $10$ epochs, the learning rate is multiplied by $0.2$ every $2$ epochs.
For continuous attention models, we use a penalty of $0.01$ in the ridge regression step. 
For multimodal continuous attention, we perform $5$ and $10$ iterations of the EM algorithm during training and testing, respectively. We set $\lambda=5$, which leads to the selection of $K^\star=1$ in $80.8\%$ of the examples, $K^\star=2$ in $12.4\%$, $K^\star=3$ in $4.4\%$, and $K^\star=4$ in $2.4\%$.

\paragraph{Results.}
The results in Table~\ref{table:results_vqa} show similar accuracies for all attention models with a slight overall advantage for the discrete attention model. Note however that the multimodal and unimodal continuous attentions use much fewer basis functions than image regions ($N \ll \overline{L}=506$).

\begin{figure*}[t]
\centering
\includegraphics[width=0.245\linewidth]{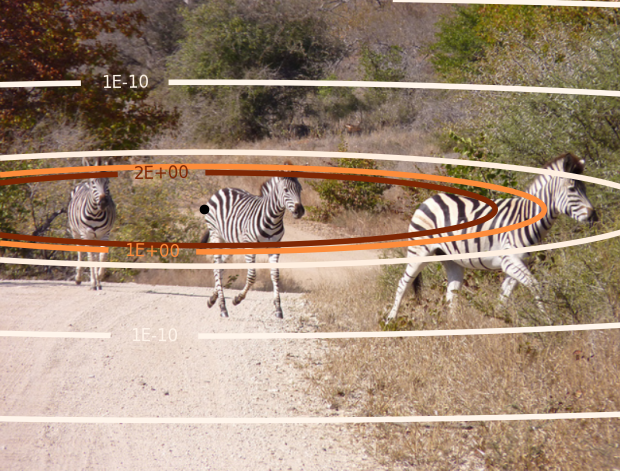}
\includegraphics[width=0.245\linewidth]{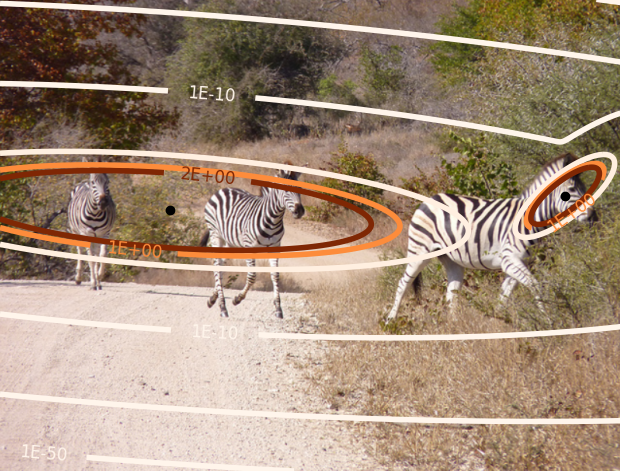}
\includegraphics[width=0.245\linewidth]{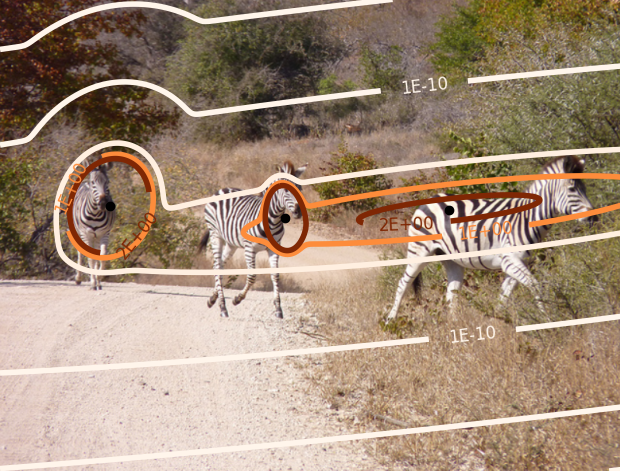}
\includegraphics[width=0.245\linewidth]{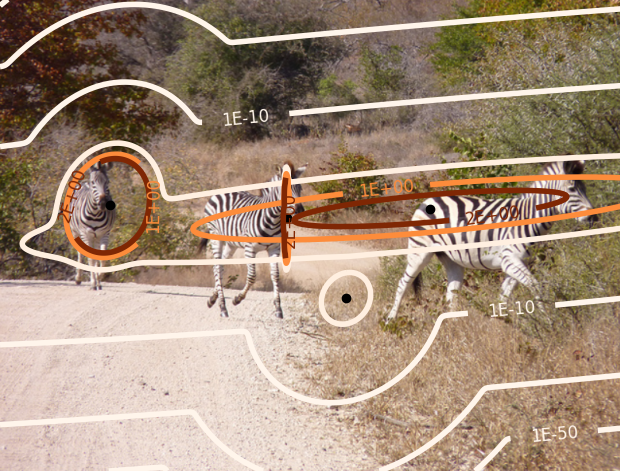}
\caption{\label{fig:horses} Attention maps generated when answering the question: \textbf{How many zebras are facing in the left direction?} Our model selection criterion chooses $K^\star=3$.
}
\end{figure*}

\begin{figure*}[t]
\centering
\includegraphics[width=0.245\linewidth]{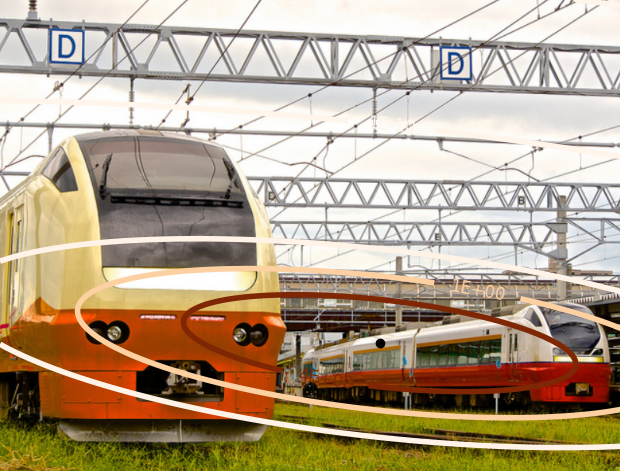}
\includegraphics[width=0.245\linewidth]{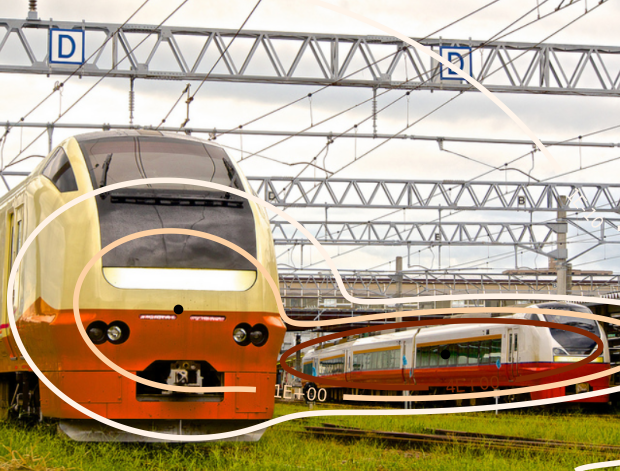}
\includegraphics[width=0.245\linewidth]{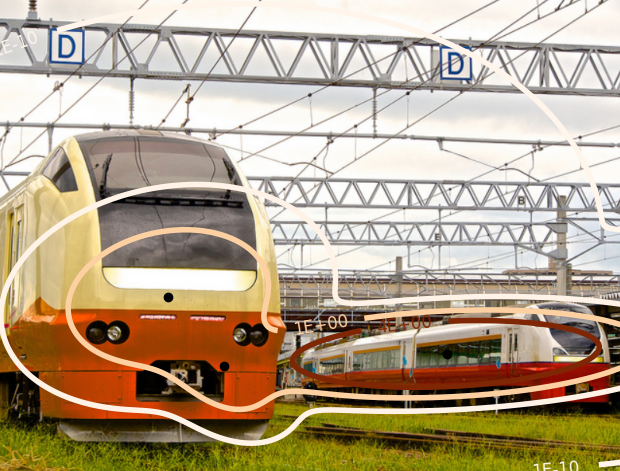}
\includegraphics[width=0.245\linewidth]{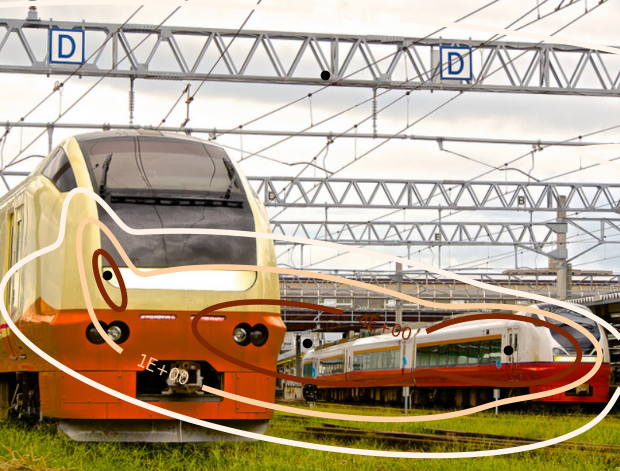}
\caption{\label{fig:trains} Attention maps generated when answering the question: \textbf{How many trains?} Our model selection criterion chooses $K^\star=2$.
}
\end{figure*}

\paragraph{Attention visualization.}
We identify two main strengths of multimodal continuous attention when compared to discrete or unimodal continuous attention.
First, previously proposed continuous attention models face difficulties in complex scenes (\eg, if there are multiple regions of interest that are far from each other), due to being limited to a single mode.
In those cases, unimodal attention ellipses become wide and less interpretable, assigning a high probability mass to a region that is not the most relevant one; or they focus on a single region and completely disregard the others.
As suggested by  Figure~\ref{fig:uni-vs-multimodal}, multimodal attention densities tend to perform considerably better in such situations, by increasing the number of components in the attention mixture and adequately setting the mean and covariance matrix of each Gaussian component.

Another interesting case is illustrated by the example in Figure~\ref{fig:toilet}. Although there is a single region of interest in the image, its complex shape confuses the non-structured and scattered discrete attention model. Besides, as a result of being overly focused, a simple Gaussian distribution is not enough to fully encompass all the relevant objects in the scene. 
By increasing the number of components in the mixture, continuous attention models become capable of more accurately segregate objects from ground, encompassing their actual shapes.

\paragraph{Comparing multimodal attention maps.}
Figures~\ref{fig:horses} and~\ref{fig:trains} illustrate how the model selection criterion \eqref{eq:model-selection-criterion-lambda} is used to estimate the number of components in the attention mixture. 
In the first example, when asked how many zebras are facing left, our attention model chooses $K^\star=3$, aligning the ellipses properly. It is interesting to see that by decreasing or increasing the number of components in the mixture, the attention map becomes less interpretable (see, for instance, that for $K=2$ there is a distribution \textit{peak} between two zebras, and for $K=4$, we can clearly identify one extra component with its mean located on the ground). 
A similar analysis can be done for the example in Figure~\ref{fig:trains}, where our model opts to use only two components.

\subsection{Human attention}
To quantitatively evaluate how interpretable different attention models are, we compare the attention distributions obtained using different models with human attention. 
For this purpose, we use the VQA-HAT dataset~\cite{vqahat} that contains human attention maps obtained through a deblurring procedure: human annotators were presented with a blurred image and a question about it, and were asked to progressively sharpen the regions of the image that help them answer the question correctly.
In order to compare the attention distributions with the human attention, we measure the Jensen-Shannon~(JS) divergence between them. This metric was proposed in~\cite{martins2020sparse} and addresses the limitations of order-based metrics like the Spearman's rank correlation used in~\cite{vqahat}, by taking into account the magnitude of the attention distributions at a given spacial location.\footnotemark

\footnotetext{The output of all the attention models is strictly dense, assigning a probability mass to every image feature. Since less relevant features are all assigned a very small positive attention value, order-based metrics are less suitable for measuring the similarity between attention distributions.
}

The results reported in Table~\ref{tab:Human Attention} show that the attention distributions obtained with multimodal continuous attention mechanisms are more similar to human attention than the ones obtained with discrete or unimodal continuous attention. These results suggest that our method is able to generate more human-interpretable attention maps.

\begin{table} 
    \begin{center}
    \begin{tabular}{lc}
        \toprule
        \sc Attention & JS divergence $\downarrow$ \\
        \midrule
        Discrete softmax & 0.64 \\
        Unimodal continuous & 0.59 \\
        Multimodal continuous & \textbf{0.54} \\
        \bottomrule
    \end{tabular}
    \end{center}
    \caption{JS divergence between attention distributions obtained with the different models and human attention.}
    \label{tab:Human Attention}
\end{table}

\begin{figure*}[t]
\centering
\includegraphics[width=0.245\linewidth]{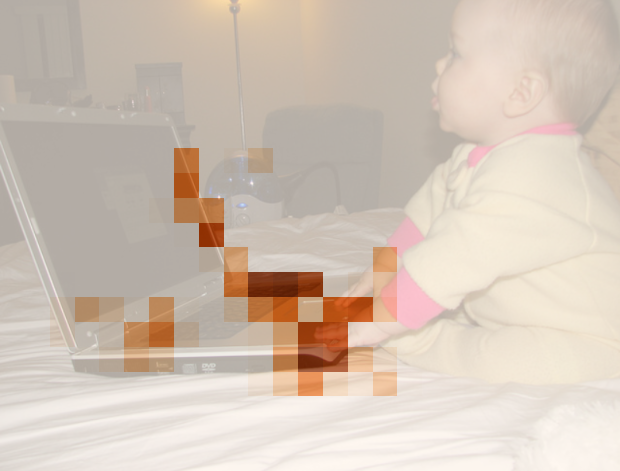}
\includegraphics[width=0.245\linewidth]{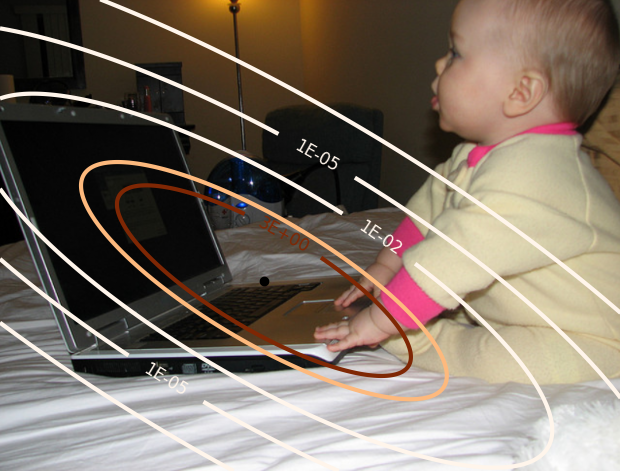}
\includegraphics[width=0.245\linewidth]{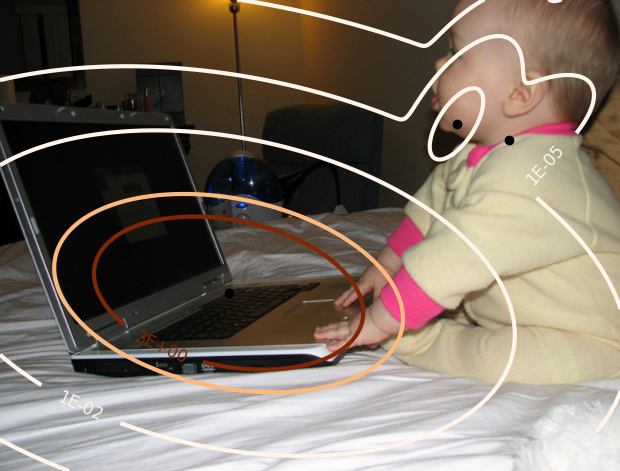}
\includegraphics[width=0.245\linewidth]{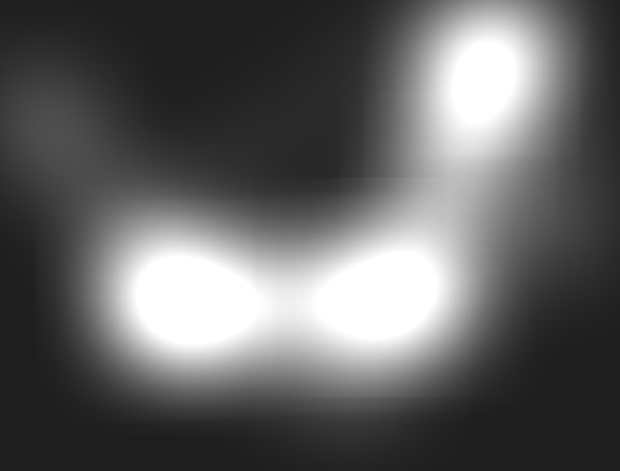}
\caption{\label{fig:human-attention-baby} Attention maps generated when answering the question: \textbf{Is the baby using the computer?} Discrete attention (JS div. $= 0.66$), unimodal continuous attention (JS div. $= 0.66$), multimodal continuous attention (JS div. $= 0.60$), human attention.
}
\end{figure*}

\begin{figure*}[t]
\centering
\includegraphics[width=0.245\linewidth]{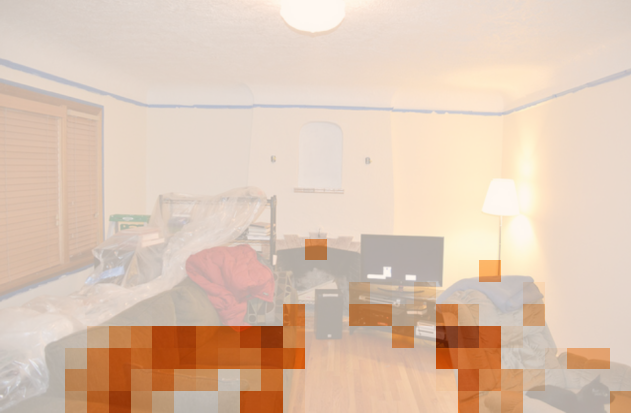}
\includegraphics[width=0.245\linewidth]{Figures/cat_unimodal_new.png}
\includegraphics[width=0.245\linewidth]{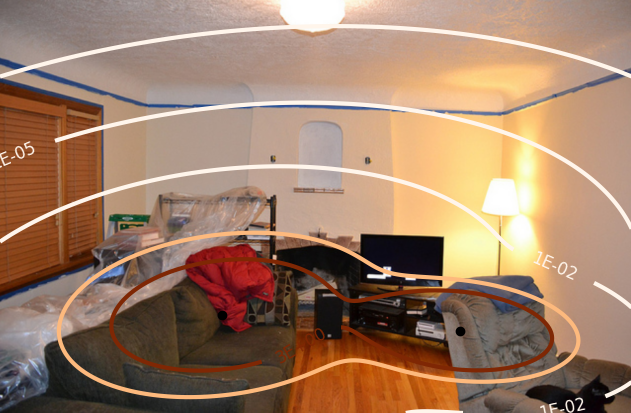}
\includegraphics[width=0.245\linewidth]{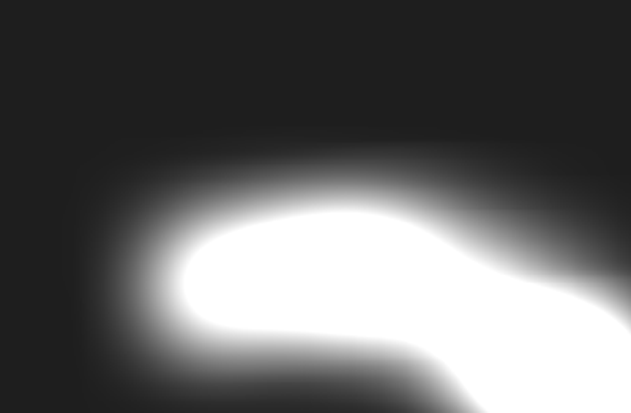}
\caption{\label{fig:human-attention-cat} Attention maps generated when answering the question: \textbf{What type of furniture is the cat sitting on?} Discrete attention (JS div. $= 0.66$), unimodal continuous attention (JS div. $= 0.56$), multimodal continuous attention (JS div.$=0.51$), human attention.
}
\end{figure*}

\begin{figure*}[ht!]
\centering
\includegraphics[width=0.245\linewidth]{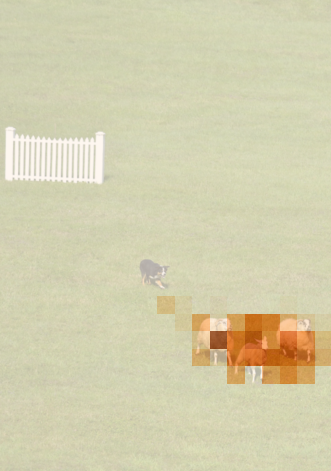} 
\includegraphics[width=0.245\linewidth]{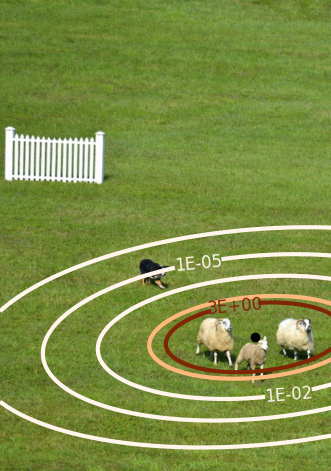}
\includegraphics[width=0.245\linewidth]{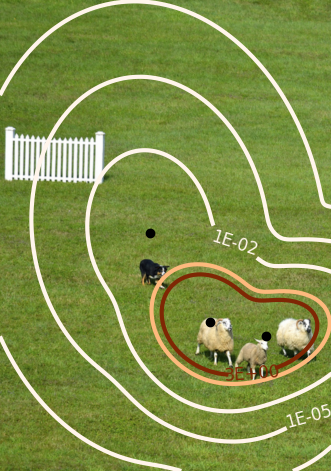}
\includegraphics[width=0.245\linewidth]{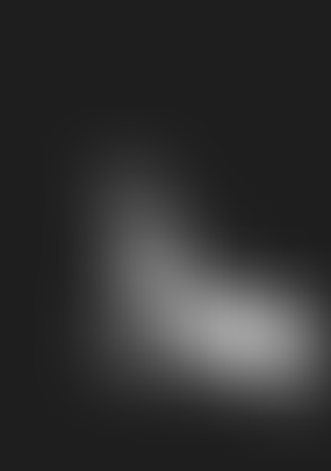}

\caption{\label{fig:human-attention-sheep} Attention maps generated when answering the question: \textbf{How many sheep are there?} Discrete attention (JS div. $= 0.68$), unimodal continuous attention (JS div. $= 0.71$), multimodal continuous attention (JS div. $= 0.68$), human attention.
}
\end{figure*}

\paragraph{Attention visualization.} 
Figures~\ref{fig:human-attention-baby}, \ref{fig:human-attention-cat} and \ref{fig:human-attention-sheep} illustrate how the attention maps generated by different attention models relate to human attention.
To answer the questions, humans sequentially look for regions in the image, until they found all the information they need. Our multimodal attention models replicate this process by identifying multiple regions of interest.

\section{Related work}
\label{section: Related work}

\paragraph{EM algorithm for weighted data.} Gebru \etal\cite{gebru2016weighteddata} proposed to incorporate the weights into the model by "\textit{observing $\bs{x}$ $w$ times}" and changing the log-likelihood function accordingly: they raise $\mathcal{N}(\bs{x};\bs{\mu},\bs{\Sigma})$ to the power $w$ and notice that $\mathcal{N}(\bs{x};\bs{\mu},\bs{\Sigma})^w \propto \mathcal{N}(\bs{x};\bs{\mu},\bs{\Sigma}/w)$, deriving a new mixture model where $w$ plays the role of precision. However, they focus on the case where the weights are treated as random variables, which is different from ours.

\paragraph{Sparse continuous attention.} 
Martins \etal\cite{ContinuousAttention2020} introduced continuous attention mechanisms for both 1D and 2D applications. 
In their work, they consider other densities besides Gaussian distributions, in particular densities with sparse support, such as truncated paraboloids, establishing a parallel with Tsallis-regularized prediction maps \cite{Tsallis1988,blondel2020learning}. 
In our work, we restrict to Gaussian densities, which are simpler and allow closed-form forward and backpropagation steps. Furthermore, mixtures of Gaussians (the multimodal extension considered in our paper) are more amenable for use in soft clustering with the EM algorithm, since they have tractable and efficient expectation and maximization steps.

\section{Conclusions and future work}
We propose new continuous attention mechanisms that produce multimodal densities in the form of mixtures of unimodal distributions (\eg a Gaussian) and show that they decompose as a linear combination of unimodal attention mechanisms, enabling tractable and efficient forward and gradient backpropagation steps (\S\ref{section: Multimodal continuous attention}). 
We use a weighted version of the Expectation-Maximization (EM) algorithm to obtain a selection of relevant regions in the image (\S\ref{section: The EM algorithm for GMMs}), and a penalized likelihood method to select the number of components in the mixture (\S\ref{section: Estimating the number of components}).
Experiments on visual question answering show that the selected regions mimic human attention more closely than previously proposed models, leading to more interpretable attention maps (\S\ref{section:Experiments}). 

There are several avenues for future research. We used mixture of Gaussians only. However, it seems interesting to consider mixtures of sparse family distributions (\eg mixtures of truncated paraboloids) in which different components may have disjoint supports. 
Another direction consists in exploring our method in other vision tasks that require learning from images and video, which could equally benefit from focusing on multiple objects simultaneously.

\section*{Acknowledgments}
This work was supported by the European Research Council (ERC StG DeepSPIN 758969),
by the P2020 program MAIA (contract 045909), and by the LARSyS - FCT Plurianual funding 2020-2023. We would like to thank Pedro Martins and Marcos Treviso for their helpful feedback. 

{\small
\bibliographystyle{ieee_fullname}
\bibliography{egbib}
}

\newpage
\appendix
\onecolumn
\begin{center}
\LARGE{\bf Supplementary Material}
\end{center}

\section{Failure cases in visual question answering}
\vspace{-0.1cm}
\begin{figure*}[!htb]
\centering
\includegraphics[width=0.33\linewidth]{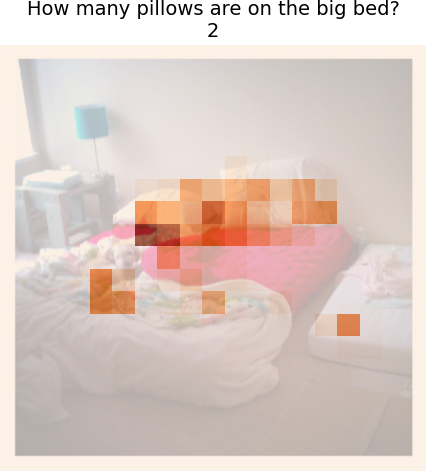}
\includegraphics[width=0.33\linewidth]{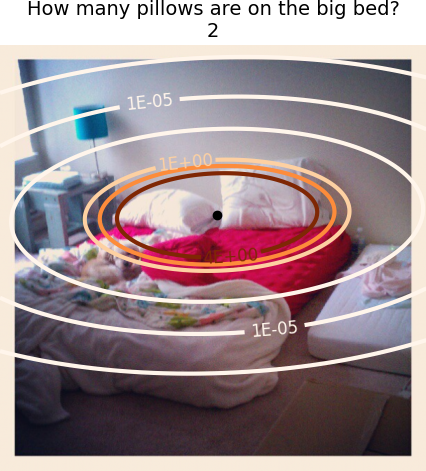}
\includegraphics[width=0.33\linewidth]{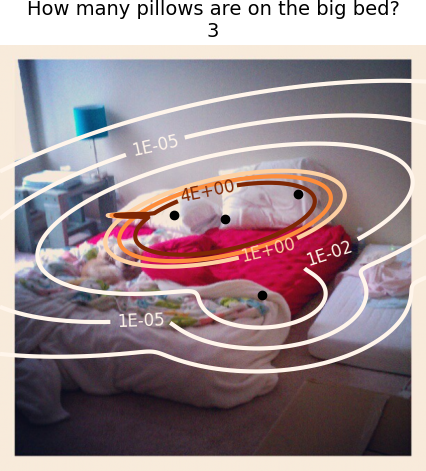}
\caption{\label{fig:pillow} Examples of attention maps in the VQA-v2 dataset. Left: discrete softmax attention. Middle: unimodal continuous attention. Right: Multimodal continuous attention (ours).
}
\end{figure*}

\vspace{-0.3cm}
\begin{figure*}[!htb]
\centering
\includegraphics[width=0.33\linewidth]{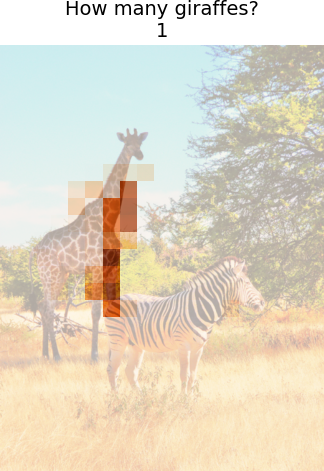}
\includegraphics[width=0.33\linewidth]{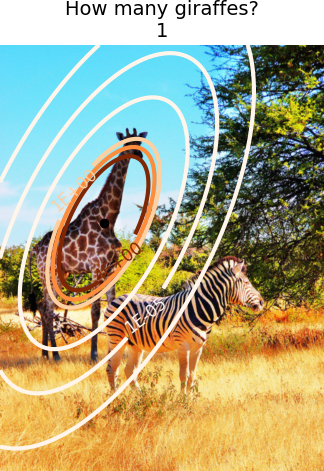}
\includegraphics[width=0.33\linewidth]{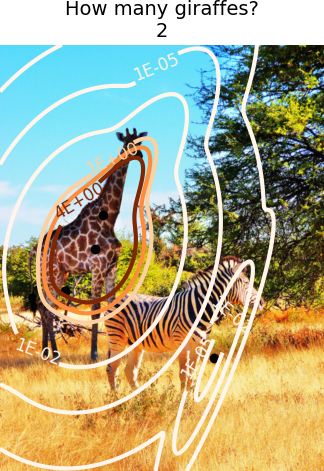}
\caption{\label{fig:giraffe} Examples of attention maps in the VQA-v2 dataset. Left: discrete softmax attention. Middle: unimodal continuous attention. Right: Multimodal continuous attention (ours).
}
\end{figure*}

\vspace{-0.1cm}
In \S\ref{section:Experiments} we presented several attention maps generated by different models and discussed the main strengths of multimodal continuous attention, when compared to discrete or unimodal continuous attention.
Although our model tends to perform considerably better in complex situations where it is possible to identify multiple regions of interest or a single region with a complex shape, there are cases in which fitting a multimodal distribution as the attention density may lead to incorrect answers. For instance, in the example in Figure~\ref{fig:pillow}, when looking for pillows on the bed, our model focuses on more than one region and possibly confuses the messy bed cover with a pillow. 
A similar situation is illustrated by the example in Figure~\ref{fig:giraffe}, where the zebra is taken as being another giraffe. 
These examples suggest that in spite of being able to generate unimodal attention maps when the relevant regions in the image are contiguous or unique, our model sometimes fails as a result of its capability of looking for more than one region in the image.

\end{document}